%% file: activellm.tex
\documentclass[11pt,a4paper]{article}
\usepackage{times,latexsym}
\usepackage{url}
\usepackage[T1]{fontenc}

\usepackage[acceptedWithA]{tacl2021v1}
\usepackage{xspace,mfirstuc,tabulary}

\newif\iftaclinstructions
\taclinstructionsfalse % AUTHORS: do NOT set this to true
\iftaclinstructions
\renewcommand{\confidential}{}
\renewcommand{\anonsubtext}{(No author info supplied here, for consistency with
TACL-submission anonymization requirements)}
\newcommand{\instr}
\fi

\iftaclpubformat % this "if" is set by the choice of options

\else

\fi

\usepackage{booktabs}       % professional-quality tables
\usepackage{graphicx}
\usepackage{amssymb}% http://ctan.org/pkg/amssymb
\usepackage{pifont}% http://ctan.org/pkg/pifont
\newcommand{\cmark}{\ding{51}}%
\newcommand{\xmark}{\ding{55}}%

\newcommand{\revised}[1]{{\color{black}#1}}
\newcommand{\revisedNEW}[1]{{\color{black}#1}}

\title{ActiveLLM: Large Language Model-based Active Learning for Textual Few-Shot Scenarios}

\author{
  Markus Bayer$^\diamond$%\Thanks{The {\em actual} contributors to this instruction
  \and
  Justin Lutz
  \and 
  Christian Reuter
  \\
  \ \\
  PEASEC
  \\
  Technical University of Darmstadt
  \\
  $^\diamond$\texttt{bayer@peasec.tu-darmstadt.de}
  \\
}

\date{}

\begin{document}
\maketitle
\begin{abstract}
    Active learning is designed to minimize annotation efforts by prioritizing instances that most enhance learning. 
    However, many active learning strategies struggle with a `cold-start' problem, needing substantial initial data to be effective.
    %This limitation often reduces their utility for pre-trained models, which already perform well in few-shot scenarios. 
    \revised{
    This limitation reduces their utility in the increasingly relevant few-shot scenarios, where the instance selection has a substantial impact.
    To address this, we introduce ActiveLLM, a novel active learning approach that leverages Large Language Models such as GPT-4, o1, Llama 3, or Mistral Large for selecting instances.
    We demonstrate that ActiveLLM significantly enhances the classification performance of BERT classifiers in few-shot scenarios, outperforming traditional active learning methods as well as improving the few-shot learning methods ADAPET, PERFECT, and SetFit.} 
    Additionally, ActiveLLM can be extended to non-few-shot scenarios, allowing for iterative selections.
    In this way, ActiveLLM can even help other active learning strategies to overcome their cold-start problem.
    Our results suggest that ActiveLLM offers a promising solution for improving model performance across various learning setups.
\end{abstract}

%\iftaclpubformat

\input{Sections/01_Introduction}
\input{Sections/02_RelatedWork}

\input{Sections/03_Method}
\input{Sections/04_Experiments}
\input{Sections/05_Conclusion}

%\iftaclpubformat

%\section{Acknowledgments}

\bibliography{activellm}
\bibliographystyle{acl_natbib}

%\iftaclpubformat
\clearpage
\onecolumn
\appendix
\revised{
\input{Sections/06_Appendix}

}

\end{document}

%% file: Sections/01_Introduction.tex
\section{Introduction}
\label{sec:introduction}

The selection of training examples significantly impacts the performance of models. 
Even Large Language Models (LLMs), such as GPT-3, show high variances based on training example selection \citep{zhang2022}.
This dependency is even more pronounced in smaller models such as BERT \citep{zhang2022}, which are sometimes preferred for their cost and resource efficiency as well as data protection benefits over APIs.
\revised{In order to combine the advantages of both the lightness of BERT-like models and the capabilities of LLMs, we explore the classical concept of Active Learning (AL) for BERT-like models, extended by LLMs such as GPT-4 \citep{openai2023GPT4}.}

\begin{figure}
    \centering
    \includegraphics[width=\linewidth]{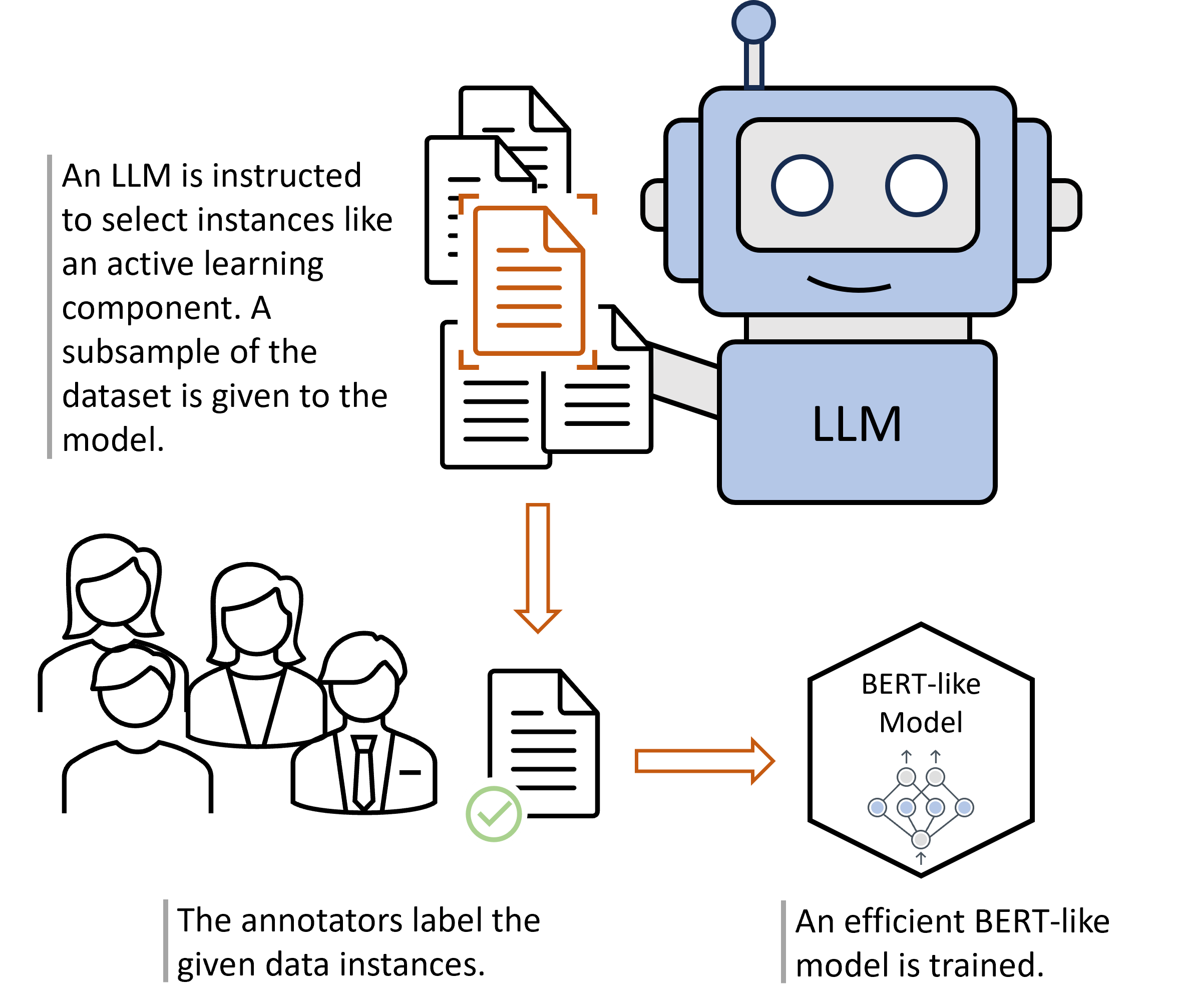}
    \caption{\revised{Depiction of ActiveLLM.}}
    \label{fig:overview}
\end{figure}

AL encompasses strategies integrated into the labeling process to select instances with a high learning impact \citep{zhang2022SurveyActiveLearning}.
Typically, a model is trained iteratively during the annotation process to query instances for labeling by the annotators (oracle). 
\revised{While AL can yield significant learning improvements with mid-sized datasets, it often encounters a `cold-start' problem \citep{chen2023MakingYourFirst}, rendering it unsuitable for low-data regimes.
The cold-start problem occurs in many AL methods because they lack sufficient data at the start of data labeling to accurately measure informativeness and select informative samples.}
For example, the uncertainty strategy selects the most uncertain instances based on an iteratively learned classifier during annotation.
However, the classifier cannot make accurate uncertainty guesses without enough data.
% \revised{Pre-trained models like BERT \citep{devlin2019BERT} alleviate this issue to some extent by leveraging prior knowledge to perform well in few-shot scenarios. 
% Nonetheless, \textcolor{red}{mixed results in research \citep{todo} show} that traditional AL strategies might require adaptation to fully exploit the capabilities of such models.
% }

\revised{Furthermore, as most AL strategies require a model to be trained iteratively, the use of pre-trained models such as BERT can result in very high delay times during annotation, making the process unusable in real-world annotation scenarios.
In so-called `model-mismatch scenarios', where the instance selection model (query model) differs from the model used for the final application (successor model), AL yields limited gains \citep{zhang2022SurveyActiveLearning}.} 

To address these challenges of few-shot and model-mismatch scenarios, we propose \mbox{ActiveLLM}, an AL method leveraging LLMs \revised{(see Figure \ref{fig:overview})}. 
\revised{\mbox{ActiveLLM} can select instances with high learning impact even without initial supervised data and requires no training during the annotation process.} 
This method serves both as a standalone AL approach and as a solution to the cold-start problem in other AL strategies.
Therefore, our contributions are as follows:

\textbf{(C1) \revisedNEW{Novel} AL Method:} We introduce a novel AL strategy using LLMs such as GPT-4, tailored for few-shot learning to overcome the cold-start problem.

\revised{\textbf{(C2) Efficient and Scalable Methodology:} Our approach decouples the query process from the successor model dependency, enhancing scalability and establishing a high degree of practicality compared to conventional AL strategies.}

\revised{\textbf{(C3) Rigorous Evaluation:} We conduct comprehensive evaluations to demonstrate that \mbox{ActiveLLM} is not constrained to a specific LLM. It exhibits superior performance compared to other AL and few-shot learning methods and can be utilized alongside these methods to achieve beneficial outcomes.} 

The code for this study will be freely available.

%The code of this study is freely available.

%(C1) A novel AL method based on LLMs suitable for few-shot scenarios and overcoming the cold-start problem.
%(C2) The method breaks the tie between query and successor model.
%(C4) Real-World Practicability (regarding time even compared to AL strategies)
%(C3) Comprehensive evaluations with different LLMs and comparisons to state-of-the-art few-shot and AL.

%% file: Sections/02_RelatedWork.tex
\section{Related Work}

\subsection{Active Learning}

The rationale behind AL is that a system selectively chooses instances to be labeled, thereby reducing the labeling effort \citep{settles2012ActiveLearning}.
In this context, annotators serve as oracles, responding to queries from the AL system regarding specific instances. % the labels of specific instances.
This system utilizes a query strategy that often hinges on measures of uncertainty or diversity. 
Traditionally, the querying is performed by the machine learning model itself, which undergoes iterative training during the labeling process.
\revised{For instance, the model might request labels for instances where its uncertainty is highest, using approaches such as Least Confidence (LC) \citep{Cohn1994LC}, Margin of Confidence \citep{Tong2001MarginOfConfidence}, or Prediction Entropy (PE) \citep{Settles2008PE}. It is then retrained with the newly annotated data.

In addition to uncertainty-based methods, diversity-based approaches also play a critical role in AL. 
These methods aim to select instances that are diverse to ensure the labeled dataset covers a wide range of feature space. 
For example, techniques like Embedding KMeans (EKM) \citep{schroeder2023smalltext, yuan2020ColdstartActive} leverage the model's ability to embed instances into a latent space. By applying clustering algorithms such as KMeans in this space, the system can identify and select the most diverse examples, thereby enriching the training set with varied and representative data points.

\subsubsection{Active Learning with Transformer Models}
\revised{While AL methods have demonstrated their utility across various traditional machine learning models, their application to transformer models remains challenging.}
The study by \citet{dor2020ActiveLearningForBERT}, among the first to explore AL with pre-trained transformer models such as BERT, shows that AL can enhance the performance of BERT classifiers, particularly in low-data scenarios and when the initial dataset is rich in relevant class instances.
However, most of the other works indicate mixed results. 
For example, \citet{jacobs2021ActiveLearningforReducingLabeling} combine uncertainty and diversity-based methods, utilizing SentenceBERT embeddings to refine the instance selection process. 
Their findings indicate that while AL reduces labeling effort, its effectiveness with BERT models is mixed, and the improvements are modest compared to older NLP models.
Studies by \citet{yuan2020ColdstartActive,schroder2022RevisitingUncertainty, seo2022ActiveLearningOnPreTrained, Griesshaber2020FinetuningBERTforLow, houlsby2011BaysianActiveLearning} contribute to a similar mixed picture of AL methods for transformer models.
In Table \ref{tab:overview}, we present an overview of works utilizing AL for transformer models, categorizing them by query strategies, tested models, and model matching/mismatching scenarios.

\textbf{Research Gap:} The most significant gap in AL research is the use of LLMs, such as GPT-4, to improve AL with BERT-like transformer models, which, to our knowledge, has not been explored previously.
}

\clearpage
\subsubsection{Model-Mismatch Scenarios}

\revised{As transformer models often require substantial computational resources and time \citep{Treviso2023EfficientNLP}, the training runtimes for those can be prohibitively long during labeling, making them impractical for real-world applications of AL.
%This is problematic for the practical application of AL, as the training runtimes for large transformer models can be prohibitively long \citep{zhang2022SurveyActiveLearning}.
\citet{schroder2022RevisitingUncertainty} noted that the incremental training of these models can be so time-consuming, that it can negate the cost savings from reduced labeling effort.
Therefore, the scope of AL has been extended to include scenarios where the querying model may differ from the final (successor) model — a phenomenon known as model mismatch — or situations where no model is used for querying.
}

%Moreover, the training of transformer models is not only a time-consuming process but requires significant resources and hyperparameter tuning. 
%The outcomes of this process can vary significantly.
%Thus, subsequent studies explore mismatch scenarios where the query model differs from the successor model.

But again many studies \citep{baldridge2004ActiveLearningTotalCost,schroder2022RevisitingUncertainty, shelmanov2021ActiveLearningSequenceTagging} report that a model mismatch between the query and successor model often leads to unsatisfactory results \citep{zhang2022SurveyActiveLearning}.
While efforts by \citet{shelmanov2021ActiveLearningSequenceTagging} and \citet{nguyen2022FAMIE} attempt to address this by using smaller, similar models for querying, there remains a dependency on the model type. 
Similarly, \citet{tsvigun2022TowardsComputationally} offer a promising approach for decoupling the query and successor models, although a distilled version is still required in the process.

\revised{\textbf{Research Gap:} In contrast to the use of smaller, resource-efficient models for querying, we propose leveraging much larger language models.
This approach may seem counterintuitive at first due to concerns about long runtimes, but as these models are capable of zero-shot learning, no training is required  \citep{Shliazhko2024FewShotLearnersMultilingual}. 
Furthermore, these models can be accessed in real-time through several free chat interfaces, requiring no machine learning expertise, making them ideal for practical applications.}

%Alternatively, the latter approach often focuses on maximizing the diversity of the selected examples and might involve techniques such as clustering.

%For a overview of AL in NLP research, including more detailed information on strategies we recommend the survey by \citet{zhang2022SurveyActiveLearning}.

\subsubsection{Cold-Start Problem}

While AL can significantly streamline the learning process by reducing labeling efforts, most strategies encounter a cold-start problem.
In the absence of sufficient initial labeled data (label seed), the AL system may struggle to make informed predictions about uncertainty or diversity \citep{yuan2020ColdstartActive}.
\revised{This presents a major challenge for low-data regimes, particularly in few-shot scenarios with very few available data instances.
% While pre-trained models generally achieve good results in low-data regimes and, therefore, the extent of the cold-start problem is less, itm   
}
%This challenge is pronounced when dealing with pre-trained models, which generally require fewer instances.
\citet{yuan2020ColdstartActive} propose ALPS, which is one of the only works that directly addresses the cold-start problem in AL with transformer models. 
ALPS leverages BERT’s pre-existing masked language modeling objective along with clustering to select instances. 
The authors hypothesize that this strategy provides more reliable confidence scores in the initial stages of model training than a classifier head.
However, the results are mixed which is also later confirmed by \citet{nguyen2022FAMIE}.

\revised{\textbf{Research Gap:} 
The cold-start problem is a major challenge in AL research, especially for few-shot scenarios. 
% When combining BERT-like models with efficient few-shot learning methods, mentioned in the following section, it stands to reason that there might be little or no improvement through AL. 
This work aims to overcome this challenge by proposing an AL strategy that needs no initial labeled data, enabling improvements even at the very beginning of the AL process.
This allows the method to also be used to address the cold-start problem of other AL methods by providing them with initial data.
}
%Moreover, these models often necessitate substantial time resources, thereby increasing the necessity for model-mismatch scenarios. 

\begin{table*}[!ht]
\centering
\resizebox{\textwidth}{!}{\begin{tabular}{lllll}%{llllll}
\textbf{Reference} &
  \textbf{Query Strategies} &
  \textbf{Model} &
  \textbf{Matching} &
  %\textbf{MCDO} &
  \textbf{Comparison To Random} \\ \hline
\citep{dor2020ActiveLearningForBERT} &
  \begin{tabular}[c]{@{}l@{}}LC, Perceptron \\ Ensemble, EGL,  \\ Core-Set, D-AL\end{tabular} & %LC (w/ and w/o \\ MCDO), Perceptron \\ Ensemble, EGL,  \\ Core-Set, D-AL\end{tabular} &
  BERT & 
  \cmark &
  %\cmark/\xmark &
  \begin{tabular}[c]{@{}l@{}}Improvements in imbalenced \\ low-data scenarios\end{tabular} \\ \\
\citep{jacobs2021ActiveLearningforReducingLabeling} &
  \begin{tabular}[c]{@{}l@{}}VR, PE, BALD, \\ Diversity Heuristics\end{tabular} &
  BERT &
  \cmark &
  %\cmark &
  Mixed results \\ \\
\citep{yuan2020ColdstartActive} &
  \begin{tabular}[c]{@{}l@{}}ALPS, PE, \\ BERT-KM BADGE\end{tabular} &
  BERT &
  \cmark &
  %\xmark &
  Mixed results \\ \\
\citep{schroder2022RevisitingUncertainty} &
  \begin{tabular}[c]{@{}l@{}}PE, Breaking Ties, \\ LC, C-AL \end{tabular} &
  \begin{tabular}[c]{@{}l@{}}BERT \\  distilRoBERTa\end{tabular} &
  \cmark &
  %\xmark &
  \begin{tabular}[c]{@{}l@{}}Small improvements in later\\ iterations\end{tabular} \\ \\
\citep{seo2022ActiveLearningOnPreTrained} &
  \begin{tabular}[c]{@{}l@{}}BATL, LC, ALPS, \\ PE, BALD, D-AL,\\ Core-Set, BADGE\end{tabular} &
  \begin{tabular}[c]{@{}l@{}}DISTRE, BERT\\ SCIBERT\end{tabular} &
  \cmark &
  %\xmark &
  Mostly improvements \\ \\
\citep{Griesshaber2020FinetuningBERTforLow} &
  BALD &
  BERT base &
  \cmark &
  %\cmark &
  Mixed results \\ \\
\citep{shelmanov2021ActiveLearningSequenceTagging} &
  \begin{tabular}[c]{@{}l@{}}MNLP, VR, BALD\end{tabular} &
  \begin{tabular}[c]{@{}l@{}}BERT base, \\ DistilBERT base, \\ ELECTRA base\end{tabular} &
  \cmark &
  %\cmark + &
  Significant improvements \\ \\
\citep{shelmanov2021ActiveLearningSequenceTagging} &
  \begin{tabular}[c]{@{}l@{}}MNLP, VR, BALD\end{tabular} &
  \begin{tabular}[c]{@{}l@{}}BERT, \\ DistilBERT, \\ ELECTRA\end{tabular} &
  \xmark &
  %\cmark + &
  \begin{tabular}[c]{@{}l@{}}Significant improvements when\\ mismatch is of the form: distilled\\ transformer (query) and normal\\  transformer (successor) \end{tabular} \\ \\
\citep{tsvigun2022TowardsComputationally} &
  \begin{tabular}[c]{@{}l@{}}LC, MNLP, \\ Mahalanobis Distance\end{tabular} &
  \begin{tabular}[c]{@{}l@{}}(Distil-)BERT, \\ (Distil-)RoBERTa,\\ (Distil-)ELECTRA,\\ XL-Net\end{tabular} &
  \xmark &
  %\xmark &
  \begin{tabular}[c]{@{}l@{}}Significant improvements when\\  successor model (from distilled\\ mismatch setting) is used for\\ pseudo-labeling an independent \\ model\end{tabular} \\ \\
\citep{nguyen2022FAMIE} &
  \begin{tabular}[c]{@{}l@{}}MNLP, BERT-KM, \\ BADGE, ALPS\end{tabular} &
  \begin{tabular}[c]{@{}l@{}}XLM-RoBERTa,\\ miniLM\end{tabular} &
  \xmark &
  %\xmark &
  \begin{tabular}[c]{@{}l@{}}Significant improvements when\\ successor model (from distilled\\ mismatch setting) is included in\\ query process by giving delayed\\  feedback to the distilled model\end{tabular} \\ \hline
 &
   &
   &
   %&
   &
  
\end{tabular}}
\caption{\revised{Overview of different works applying AL to transformer models. "Model" describes the successor model trained from the AL process. "Matching" indicates if the successor model matches the query model.}}
\label{tab:overview}
\end{table*}

\subsection{Few-Shot Learning}

\revised{Few-shot learning refers to methods for scenarios where models need to learn from a limited number of training instances. 
%This is particularly challenging in fields where acquiring large datasets is difficult or expensive.
These scenarios are becoming increasingly relevant due to their importance in real-world problems \citep{Yu2020FewShotImportance} and the increasing capabilities of deep learning models.}
While LLMs such as GPT-3 inherently have robust few-shot or zero-shot capabilities through in-context learning, our focus is on enhancing few-shot learning in smaller models such as BERT.
\revised{With specific few-shot learning methods like ADAPET \citep{tam2021ImprovingandSimplifying} these smaller language models can attain similar or even better few-shot performance than GPT-3.}

One approach to this is to formulate tasks as cloze-style tests, where certain words in a text instance are missing and the language model is employed to fill in the blanks.  
%Task instances are often rephrased into questions so that the answer word to fill corresponds to the label.
\revised{This way, no classifier needs to be trained on top of the language model, leading to a more effective utilization of the language models \citep{gao2021MakingPretrainedLanguage}.
Such an approach is exemplified by the ADAPET method \citep{tam2021ImprovingandSimplifying}, as well as in the works of \citet{gao2021MakingPretrainedLanguage, zhang2022DifferentiablePrompt, schick2021ExploitingCloze}.
PERFECT, introduced by \citet{Mahabadi2022PERFECT}, eliminates the need to manually formulate such cloze tasks by incorporating task-specific adapters and specialized label embeddings.
Similarly, in SetFit by \citet{tunstall2022EfficientFewShot}, no cloze tasks are required, as a sentence transformer \citep{reimers2019SentenceBERT} is fine-tuned on the available training data.
The instances are then encoded with the resulting model and a regular classification head is trained on them.
%At inference, the instances are encoded by the fine-tuned sentence transformer and then classified by the trained classification head.

\revisedNEW{While not few-shot methods per se, transfer learning and data augmentation are often employed to address few-shot scenarios.
Transfer learning allows models to leverage knowledge from related domains \citep{ruder2019transfer}, providing a strong foundation in few-shot scenarios. 
Data augmentation, particularly using LLMs \citep{ding2024data, wang2022contrastive, xullm}, has been used to generate high-quality synthetic data in low-data regimes \citep{zheng2023self}.
Similarly, in this study we investigate whether AL can be useful for few-shot learning, especially by overcoming the cold-start problem.
}

\textbf{Research Gap:} 
\revisedNEW{In general, LLMs dominate the machine learning field, in particular in few-shot learning, but there are numerous reasons why researchers and practitioners still rely on smaller BERT-like models, especially for inference, where concerns such as memory limitations, API costs, and privacy often preclude the use of LLMs \citep{wang2024comprehensivesurveysmalllanguage, samuel2024BertsGen, bosley2023we}.
To the best of our knowledge, our approach is the first to leverage the capabilities of LLMs to support instance selection for training smaller BERT-like models. 
In doing so, we combine the flexibility and efficiency of smaller models with the strategic guidance of LLMs.}
%To the best of our knowledge, we are the first to combine the strengths of those LLMs with the efficiency and flexibility of BERT-like models by leveraging LLMs to select instances for labeling, which are then used to train BERT-like models.

While \citet{dor2020ActiveLearningForBERT} and \citet{Griesshaber2020FinetuningBERTforLow} also assess the performance of AL with BERT in low-resource scenarios, their datasets are still considerably larger than those in few-shot scenarios.
However, in scenarios with very few instances, example selection has the greatest impact.
In this work, we aim to improve state-of-the-art few-shot learning methods with our AL approach.
}

%% file: Sections/03_Method.tex
\section{Method}
\label{sec:method}

\begin{figure*}[htp]
  \centering
  \includegraphics[width=0.8\textwidth]{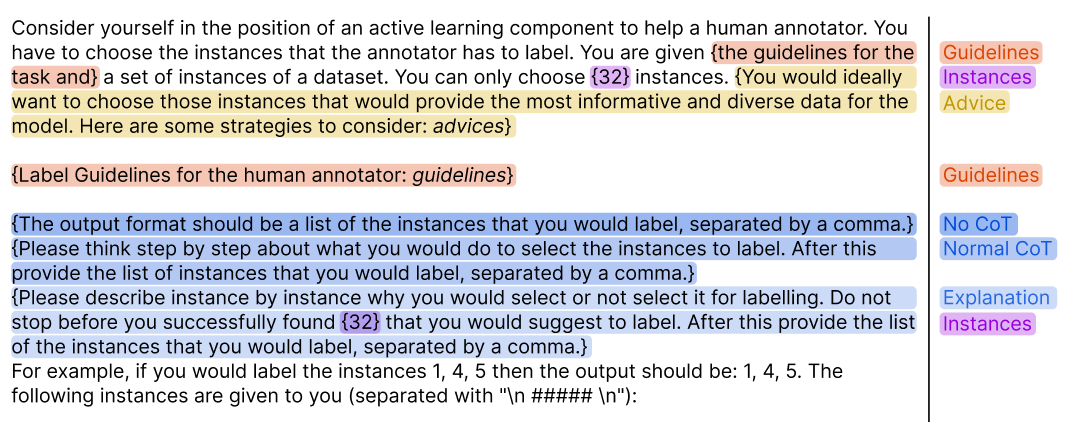}
  \caption{Prompt design of ActiveLLM in few-shot learning mode.}
  \label{fig:prompt_engineering}
\end{figure*}

ActiveLLM is a pool-based sampling method that operates in batch mode, meaning that it selects a subset from a pool of unlabeled data for querying an oracle.
%This method is particularly suited for scenarios involving model mismatch. 
It employs instruction-tuned LLMs as query models, while allowing the choice of a successor model to be independent of these models.
\revised{A demonstration of the process is given in Figure \ref{fig:overview}, where the LLM is instructed to select instances, which are then labeled and used to train a BERT-like model, allowing any such model to be trained.}
%ActiveLLM does not train the instruction-tuned LLMs during the AL process, enabling direct application to the unlabeled dataset without encountering the cold-start problem.

The design of the prompts is critical for achieving optimal results with instruction-tuned LLMs. 
To this end, we have crafted detailed prompts in various configurations to determine their effectiveness in this context. 
\revised{We differentiate between two modes of ActiveLLM. The first addresses few-shot scenarios, where ActiveLLM is executed only once on the dataset. The second is suitable for general scenarios involving iterative querying, which also incorporates feedback from previous iterations.}

\subsection{Few-Shot Learning Mode} 

The prompts are constructed so that the language model receives a description of AL, including exact details of its tasks, while also keeping the context length small, as this is a limiting factor for current models \citep{Liu2024LostInTheMiddle}.
Given a specific task, ActiveLLM creates a prompt consisting of AL role allocation, instructions on the selection process, a description of how to format the output, followed by a batch of unlabeled instances. 
%The AL role allocation positions the language model as an AL component, explaining that it should select instances for a human annotator to label. 
%Then, the instructions on the selection process specify how many instances it should select. 
%The output format should be a simple list of the indices of the selected instances.

An excerpt of the prompt design is illustrated in Figure~\ref{fig:prompt_engineering}.
The parameters displayed on the right indicate certain text phrase constellations which can be included, excluded, or adjusted:

\textit{Guidelines:} This includes a phrase indicating that the model is given not only a set of instances but also guidelines for the task.
%    Before listing the instances, the guidelines are included.
    
\textit{Instances:} This parameter determines how many instances the model should label. 
%While in the few-shot experiments this parameter is set to 32, 
We test different values in Section \ref{sec:selection_size}.
    
\textit{Advice:} Initial experiments showed that the LLMs often respond with explaining general strategies for choosing instances in an AL manner.
Hence, we included the parameter `advice,' which incorporates these strategies.% (in the figure, this is abbreviated, but we included the full prompt in Appendix \ref{sec:prompt_design_feedback_mode} Figure \ref{fig:prompt_engineering_extended})

\textit{No CoT/CoT/Explanation:} One of the most common and beneficial prompt engineering strategies is the use of Chain of Thought (CoT) prompting \citep{wei2022ChainOfThought}. 
We experiment with no CoT, `normal' CoT ("think step by step"), and the task to explain the thoughts on each instance.
%In the case of the explanation for each instance, we also reiterated the `instances' parameter, as we noticed due to very long answers that models often forget how many instances they should select.

Finally, a batch of unlabeled instances is appended to the prompt. We hypothesize that, contrary to common AL strategies, large batches or even all instances do not need to be evaluated by the query model to find instances with high learning impact. 
We test different values of this parameter in Section \ref{sec:presented_batch_size}. 

\subsection{Iterated Querying Mode}

In the second variant, we design ActiveLLM similar to other AL strategies, allowing the LLM to query new instances repeatedly.
For this, it is sensible to include data that was selected in previous iterations.
As the prompt should not become too large, we consider three additional parameters \revised{(see Appendix \ref{sec:prompt_design_feedback_mode} and Figure \ref{fig:prompt_engineering_extended} for the full prompt)}:
%These instances must be incorporated into the prompt in such a way that the prompt does not become too large. 
%Therefore, we consider three additional prompt parameters:

\textit{No Recap:} Indicating the baseline, where no instances from previous queries are recalled.

\textit{Recap:} The instances from previous queries are directly included in the prompt.

\textit{Index Recap:} To reduce the context size, the indices of the instances from previous queries are included in the prompt.

For the latter two scenarios, we included instructions in the prompt to indicate that instances from previous iterations are incorporated. 
In both cases, we did not include the annotated labels, as our experiments showed that they were generally disregarded by the language models.
\revised{An example of a specific prompt and LLM answer is given in Appendix Figure \ref{fig:input_prompt}.}

%The corresponding prompts and further details on how these designs are implemented are provided in Appendix \ref{sec:prompt_design_feedback_mode}.

%% file: Sections/04_Experiments.tex
\section{Experiments}

\revised{
\subsection{Description}

Our experimentation perspective with ActiveLLM focuses on addressing the following questions:

\begin{enumerate}
    \item How should the prompt be designed? (Section \ref{sec:prompt_engineering}) 
    \item Can the chosen prompt be applied to other models and datasets? (Section \ref{sec:main_experiments}) 
    \item How does ActiveLLM compare to other AL strategies? (Section \ref{sec:comparison})
    \item Can ActiveLLM improve state-of-the-art few-shot learning methods? (Section \ref{sec:few_shot_learning})
    \item How does the method perform in non-few-shot-scenarios? (Section \ref{sec:iterated_querying}) 
    \item Is ActiveLLM capable of overcoming the cold-start problem in other AL strategies? (Section \ref{sec:overcoming_coldstart})
\end{enumerate}

As is common in research, our experiments simulate the AL cycle. 
To do this, we take a subset of the training set (labels removed) for the respective task and allow the LLM to select a certain number of instances based on the prompt given in Section \ref{sec:method}. 
These instances are then assigned the true labels as if they had been annotated by a perfect annotator.
}

\subsubsection{Datasets}

For the initial prompt engineering experiments, we are interested in a less common dataset, as an LLM might be less biased by data leakage.
We chose the Specialized CTI dataset \citep{bayer2023MultiLevelFineTuning}, where Twitter posts are classified according to their relevance to specific CTI events \revised{(further details in Appendix \ref{sec:dataset_appendix})}.
%Detailed dataset specifications are available in Appendix \ref{sec:dataset_appendix} Table \ref{tab:dataset_splits}. 
In the general applicability tests, we used the best-performing prompt from the prompt engineering phase on the commonly-used GLUE benchmark. 
For comparison with the few-shot learning method SetFit, we use the AGNews dataset \citep{zhang2015characterlevel}, for which an adaptation of the method is available.
The SST-2 dataset \citep{socher2013RecursiveDeepModels}, previously used in the general applicability section, is also employed in the non-few-shot experiments.
\revised{To measure the performance we use the test sets for CTI and AGNews, and the validation sets for GLUE.}
\revisedNEW{Unless otherwise specified (see Sections \ref{sec:selection_size} and \ref{sec:non-few-shot}), we select 32 instances per task, either via AL or random sampling for the baseline, which represents a common few-shot learning setting \citep{tam2021ImprovingandSimplifying, Schick2021NotJustSize, Mahabadi2022PERFECT}.} 

During the prompt engineering experiments, we aim to minimize random factors by generating five different randomizations of the CTI task and applying both the baseline and each prompt to these variations.
For each evaluation, we select five different seeds for training the successor model, with the reported results for the baseline and each prompt being the average score from 25 runs (5 dataset randomizations times 5 trained models). 
For subsequent experiments, we create one randomization of the datasets but continue to train five models and provide the average score.

\subsubsection{Models and Hyperparameters}

\revised{In our experiments, we primarily employed GPT-4 \citep{openai2023GPT4} as the query model for ActiveLLM, as it achieved top results at the time the main experiments were conducted.
However, in Section \ref{sec:main_experiments} we also experiment with other LLMs: o1 \citep{OpenAI2024o1}, GPT-4o, GPT-3.5 \citep{brown2020GPT3}, Llama 3 70B, Gemini-Ultra 1.0 \citep{google2023Gemini}, Mistral Large, and Mixtral 8x7B \citep{jiang2024Mixtral} (ActiveLLM is renamed according to the used LLM - e.g. GPT-4: ActiveGPT4).
To underscore the practicality of our method, we utilize chat versions of the LLMs to simulate realistic interactions within chat environments (further details in Appendix \ref{sec:llm_details}).

As a successor model for the selected instances of ActiveLLM, we chose BERT \citep{devlin2019BERT} and RoBERTa \citep{Liu2019RoBERTa}.}
While foundational works such as from \citet{devlin2019BERT} typically train BERT-like models for only 3-5 epochs, other studies \citep{gao2021MakingPretrainedLanguage, mosbach2021StabilityFineTuning} suggest that higher epoch counts are necessary for stable and optimal performance in low-data regimes. 
Consequently, we have opted for 25 epochs for our experiments. 
Otherwise, we adhere to the default Huggingface transformer parameters, with a learning rate of 1e-3, a weight decay of 0.01, and a dropout probability of 0.1.

\revisedNEW{
For comparison, we include in each experiment a random sampling baseline where the first instances from a shuffled dataset are selected.
In addition, we compare ActiveLLM with the most commonly used AL strategies for transformers identified in our literature overview (see Table~\ref{tab:overview}): LC, BALD, EKM, and PE.
The strategies are implemented in a model-matching scenario (i.e., training the same query and successor model), can query the entire dataset of each task, and select 5 instances until the selection of 32 is reached.
ActiveLLM, on the other hand, is implemented in a model-mismatch scenario, can only query a subset of the full training set, and selects the 32 instances in one query (unless otherwise specified).
To implement the common AL strategies, we use the small-text library \citep{schroeder2023smalltext}, which provides a unified framework for pool-based AL with support for transformer-based models.
}

\subsection{Prompt Engineering Experiments}
\label{sec:prompt_engineering}

\textcolor{gray}{\textit{- How should the prompt be designed?}}

\revised{To answer the first question, we experiment with different configurations of ActiveLLM, such as whether to include AL strategy advices or guidelines, as shown in Figure \ref{fig:prompt_engineering}. 
We then evaluate the effects of varying batch sizes for the unlabeled data used in the prompts. 
Finally, we compare different selection sizes, i.e. the number of instances the LLM should select.
While we want to find the most appropriate prompt, our goal is not to engineer every detail of it, which is left open for future work and practitioners.}

\subsubsection{Configuration Results}
\label{sec:configuration_results}

\begin{table}[t]
\centering
\begin{tabular}{llllll}
          & A & CoT & G & F1 {\footnotesize(SD)}              \\ \hline
Baseline  & -      & -   & -            & 0.6781 {\scriptsize(0.029)}         \\ \hline
Prompt A1 & \cmark      & \xmark        & \xmark          & 0.6522 {\scriptsize(0.100)}        \\
Prompt A2 & \cmark      & \cmark        & \xmark          & 0.6951 {\scriptsize(0.102)}          \\
Prompt A3 & \cmark      & \cmark+       & \xmark          & 0.6876 {\scriptsize(0.108)}         \\
Prompt B1 & \xmark      & \xmark        & \xmark          & 0.7135 {\scriptsize(0.060)}         \\
Prompt B2 & \xmark      & \cmark        & \xmark          & \textbf{0.7214} {\scriptsize(0.053)} \\
Prompt B3 & \xmark      & \cmark+       & \xmark          & 0.6271 {\scriptsize(0.139)}          \\
Prompt C2  & \xmark      & \cmark       & \cmark          & \textbf{0.7329} {\scriptsize(0.053)} \\ \hline
\end{tabular}
\caption{\revised{Prompt engineering results of the baseline and ActiveGPT4 on the CTI dataset \revisedNEW{sampling 32 instances}. The results are averaged over 25 runs (5 random dataset and 5 random model initializations). A (advice), G (guidelines), C (no advice, but guidelines), 1 (no CoT), 2 (CoT: step-by-step), 3 (CoT+: explain each instance).}} %and the standard deviation is given in parentheses. The naming is given as follows: A (advice), B (no advice), C (no advice, but guidelines), 1 (no CoT), 2 (step-by-step CoT), and 3 (explain each instance CoT).
\label{tab:prompt_engineering}
\end{table}

Table \ref{tab:prompt_engineering} presents the results of various prompt configurations.
For these, we use a batch of 300 presented unlabeled instances.
Prompt configurations A and B differ in whether the advice on how to select the instances is already included in the prompt.
We observed that GPT-4 tends to reiterate the advice even when it is provided in the prompt. 
Furthermore, the results suggest that it is preferable not to include the advice in the prompt, likely due to the increased context size when the advice is repeated in both the prompt and the response.

We also test the three CoT variants: A1 and B1, where there is no CoT; A2 and B2, which use the standard CoT approach ("think step by step"); and A3 and B3, which instruct to include explanations for each instance.
The common "think step by step" instruction yielded the best results. 
This might be because this approach allows GPT more reasoning in structured steps, while not increasing the context size too much. 
%The explanations for each instance, on the other hand, are probably not helpful due to the increased context length, as explanations for each instance significantly lengthen the responses.

Finally, we experimented with including labeling guidelines in the prompt. 
\revised{As anticipated, this prompt configuration produced the best results, likely because it enabled the model to better differentiate between instances.
Although labeling guidelines are often created prior to the labeling process, they are rarely included in the dataset for common tasks.
While the guidelines are available for the CTI task, its inclusion in the prompt should be seen as an exploration of what can be achieved rather than the default case for ActiveLLM.
Therefore, in future evaluations, we will utilize ActiveGPT4 with prompt B2, i.e., no guidelines or additional advice on AL, only employing the simple "think step by step" CoT prompting.}

\subsubsection{Presented Batch Size Results}
\label{sec:presented_batch_size}

\begin{figure}[]
  \centering
  \includegraphics[width=0.47\textwidth]{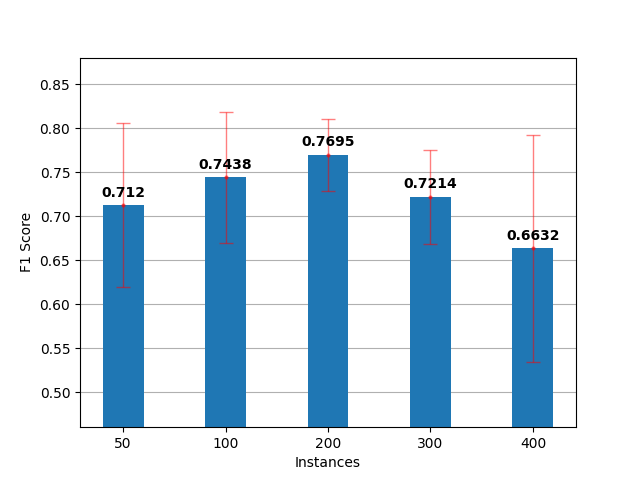}
  \caption{Varying size of the batch of unlabeled instances from which ActiveGPT4 can \revisedNEW{select 32 instances}. All results are averaged over 5 runs on the CTI dataset (F1).}
  \label{fig:presented_examples}
\end{figure}

In the subsequent experiment, we explore varying batch sizes of unlabeled instances presented to ActiveGPT4. 
Figure \ref{fig:presented_examples} displays the results for 50, 100, 200, 300, and 400 examples. There is a noticeable trade-off between the context length and the number of presented examples. While a larger pool of examples generally allows for a more diverse set, an increase in context length tends to degrade the model’s performance. 
It can be observed that the optimal prompt with 300 examples, as used in the previous section, could be significantly enhanced by reducing the size to 200.
%With 400 examples, GPT-4 occasionally indicated an inability to select instances due to the overwhelming quantity. 
Consequently, for the primary experiments in Section \ref{sec:main_experiments}, we limited the number of instances presented to the model to 200, contingent on the model's capacity to handle such a context size.

\subsubsection{Selection Size Results}
\label{sec:selection_size}

\begin{figure}[]
  \centering
  \includegraphics[width=0.47\textwidth]{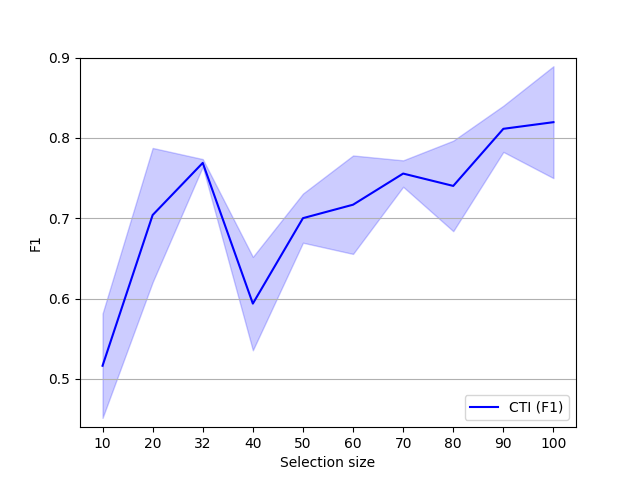}
  \caption{\revised{Varying selection size of examples to be selected by ActiveGPT4. All results (F1) are averaged over 5 runs on the CTI dataset.}}
  \label{fig:selection_size_cti}
\end{figure}

In our final experiment on prompt engineering, we investigate the impact of the selection size—how many instances the model should choose—on the outcomes.
\revised{Many research works \citep{tam2021ImprovingandSimplifying, Schick2021NotJustSize, Mahabadi2022PERFECT} typically use 32 examples for few-shot scenarios.} 
We evaluate whether smaller or larger sizes might be more effective for ActiveGPT4.
\revised{Figure \ref{fig:selection_size_cti} illustrates varying selection sizes for the CTI dataset.}
\revised{Selecting 32 examples per query appears to be a sensible choice. Notably, performance declines with up to 90 examples, suggesting that the model may struggle to process such a large number of examples effectively.
However, performance increases beyond this point, as expected, because a larger training set is inherently beneficial.}

\revised{As part of the following experiments, we also tested the selection size on the SST-2 dataset.
The results are equivalent and can be found in Appendix Figure \ref{fig:selection_size_sst2}.}

\subsection{General Applicability}
\label{sec:main_experiments}

\textcolor{gray}{\textit{- Can the chosen prompt be applied to other models and datasets?}}

\begin{table*}[h!]
{\setlength{\tabcolsep}{4pt}
\centering

\begin{tabular}{lllllllllll}
 &  & QNLI & QQP & RTE & SST2 & WNLI & MNLI-(m/mm) & MRPC & COLA %& STSB & Avg 
  \\ \hline
& Baseline & 55.11 & 63.67 & 50.83 & 56.03 & 35.77 & 34.92/35.01 & 67.55 & 63.89 & %0 & 51.42 
\\ \hline 
& GPT-4 & 58.94 $\uparrow$ & 63.99 $\uparrow$ & 53.57 $\uparrow$ & \textbf{73.14} $\uparrow$ & 40.28 $\uparrow$ & 35.45/35.14 $\uparrow$ & 68.33 $\uparrow$ & 64.70 $\uparrow$ %& 0 & 54.84 
\\ 
\rotatebox[origin=c]{90}{200} & GPT-4o & \textbf{62.76} $\uparrow$ & 63.65 \textcolor{red}{$\downarrow$} & 48.88 \textcolor{red}{$\downarrow$} & 66.31 $\uparrow$ & 41.13 $\uparrow$ & 32.23/32.80 \textcolor{red}{$\downarrow$}& 69.75 $\uparrow$ & 65.41 $\uparrow$ %& 0 & \textbf{??} 
\\
& o1 & 62.10 $\uparrow$ & 64.00 $\uparrow$ & 54.01 $\uparrow$ & 59.08 $\uparrow$ & 39.43 $\uparrow$ & 33.52/33.94 \textcolor{red}{$\downarrow$} & 63.87 \textcolor{red}{$\downarrow$} & \textbf{69.91} $\uparrow$ %& 0 & 54.84 
\\ 
& Mistral Large & 60.23 $\uparrow$ & 65.26 $\uparrow$ & \textbf{55.60} $\uparrow$ & 60.64 $\uparrow$ & \textbf{52.96} $\uparrow$ & 36.59/35.92 $\uparrow$ & 67.06 \textcolor{red}{$\downarrow$} & 67.44 $\uparrow$ %& 0 & \textbf{55.74} 
\\\hline
 & Llama 3 70B & 56.57 $\uparrow$ & 65.82 $\uparrow$ & 52.27 $\uparrow$ & 65.71 $\uparrow$ & 40.28 $\uparrow$ & 37.19/36.68 $\uparrow$ & 66.86 \textcolor{red}{$\downarrow$} & 68.15 $\uparrow$ %& 0 & 55.44 
\\
& Mistral Large & 60.68 $\uparrow$ & 64.82 $\uparrow$ & 49.48 \textcolor{red}{$\downarrow$} & 66.86 $\uparrow$ & 46.48 $\uparrow$ & 37.41/36.58 $\uparrow$ & 69.07 $\uparrow$ & 67.56 $\uparrow$ %& 0 & 55.44 
\\
\rotatebox[origin=c]{90}{100} & GPT-3.5 & 55.11 \hspace*{0.1em}-\hspace*{0.1em} & \textbf{65.88} $\uparrow$ & 55.09 $\uparrow$ & 61.15 $\uparrow$ & 51.55 $\uparrow$ & \textbf{37.44/37.62} $\uparrow$ & 67.70 $\uparrow$ & 68.63 $\uparrow$ %& 0 & 55.57 
\\ 
& Gemini-Ultra & 57.60 $\uparrow$ & 61.18 \textcolor{red}{$\downarrow$} & 54.58 $\uparrow$ & 49.08 \textcolor{red}{$\downarrow$} & 43.10 $\uparrow$ & 35.74/35.77 $\uparrow$ & \textbf{69.95} $\uparrow$ & 69.22 $\uparrow$ %& 0 & 52.91 
\\ 
& Mixtral 8x7B & 61.13 $\uparrow$ & 65.27 $\uparrow$ & 50.97 $\uparrow$ & 68.00 $\uparrow$ & 41.41 $\uparrow$ & 33.62/33.75 \textcolor{red}{$\downarrow$}& 68.14 $\uparrow$ & 66.96 $\uparrow$ %& 0 & 54.36 
\\ 
\hline
%& Classic AL & & & & & & & & &%&&
%\\
& LC & 56.76 $\uparrow$ & 58.65 \textcolor{red}{$\downarrow$} & 54.44 $\uparrow$ & 56.86 $\uparrow$ & 42.82 $\uparrow$ & 36.48/36.18 $\uparrow$ & 66.96 \textcolor{red}{$\downarrow$} & 69.26 $\uparrow$ &%&&
\\
\rotatebox[origin=c]{90}{AL} & BALD & 59.91 $\uparrow$ & n.a.*  & 52.13 $\uparrow$ & 69.17 $\uparrow$ & 44.79 $\uparrow$ & 36.14/36.34 $\uparrow$ & 68.53 $\uparrow$ & 64.24 $\uparrow$ &%&&
\\
& EKM & 52.44 \textcolor{red}{$\downarrow$} & 59.62 \textcolor{red}{$\downarrow$} & 47.65 \textcolor{red}{$\downarrow$} & 69.77 $\uparrow$ & 46.49 $\uparrow$ & 32.87/32.75 \textcolor{red}{$\downarrow$} & 61.08 \textcolor{red}{$\downarrow$} & 66.60 $\uparrow$ &%&&
\\
& PE & 53.75 \textcolor{red}{$\downarrow$} & 58.65 \textcolor{red}{$\downarrow$} & 54.44 $\uparrow$ & 56.86 $\uparrow$ & 42.82 $\uparrow$ & 37.11/37.20 $\uparrow$ & 66.96 \textcolor{red}{$\downarrow$} & 69.26 $\uparrow$ &%&&
\\\hline
\end{tabular}}
\caption{\revised{BERT-base few-shot results (32 instances) of ActiveLLM with different LLMs and of other AL strategies (LC, BALD, EKM, and PE) on GLUE tasks. `Active 200/100' describes the size of the batch of unlabeled instances. All results are averaged over 5 runs (accuracy). Standard deviations are given in Appendix Table \ref{tab:main_few_shot_results_std}. \\ **No result available after 48 hours of execution.}} %Only GPT-4, GPT-4o and Mistral Large are capable of processing 200 instances. All results are averaged over 5 runs (accuracy).}
\label{tab:main_few_shot_results}
\end{table*}

In this section, we aim to assess how ActiveLLM equipped with various LLMs and the prompt configuration B2 from Section \ref{sec:configuration_results} performs on the more common datasets of the GLUE benchmark. 
%Unlike previous experiments, we only sample one  dataset for each task but continue to average the performance across five training runs.

The results of this experiment are given in Table \ref{tab:main_few_shot_results}.
\revised{ActiveGPT4, ActiveGPT4o, Activeo1 and ActiveMistralLarge are given 200 unlabeled instances as identified in the batch size experiments (Section \ref{sec:presented_batch_size}).}
Llama 3, GPT-3.5, Gemini-Ultra and Mixtral can not process such extensive prompts and are therefore given 100 instances. 
\revised{All LLMs show that they are very good query models for ActiveLLM, outperforming the baseline in most cases. 
However, we recommend using ActiveLLM with GPT-4 or GPT-3.5, as they consistently improve results across all tasks.
%Interestingly, the o1 model, despite detailing a plan for selecting instances, often proceeded by explicitly stating that it would select instances at random, as no specific strategy was provided in the instructions.
%Therefore, the model produces more inconsistent results than GPT-4 and GPT-3.5.
\revisedNEW{An interesting finding} is that Mistral Large with 200 examples performed worse than its 100 example counterpart in five out of eight tasks.
The tasks where Mistral Large performs better with 200 instances are those with generally shorter text length, suggesting that choosing a batch size based on token length may be better than a static batch size. 
This would also explain the variation in improvements across tasks and models, as some tasks involve many more tokens per instance, and some models perform better with larger prompts.} 
\revisedNEW{In addition, the differences between the models may be due to the fact that the prompt design phase was carried out using GPT-4.
Other models may work better and give more consistent results with a prompt design specifically for them.}
The most significant improvements can be observed with ActiveMistralLarge (200) and ActiveGPT4 on WNLI and SST-2, respectively, achieving improvements of 17.19 and 17.11 percentage points over the baseline.

\revised{Overall, ActiveLLM, utilizing GPT-4 and GPT-3.5, proves to be a robust approach for enhancing performance in few-shot scenarios across all tasks.
While it may even produce better results with other LLMs, the improvements may be inconsistent and should be compared to a baseline.

\revisedNEW{
To verify that our results are not specific to BERT, we repeated the experiments from Table \ref{tab:main_few_shot_results} using RoBERTa.
The results, shown in Appendix Table \ref{tab:main_few_shot_results_roberta}, confirm that ActiveLLM also improves performance on a different BERT-based architecture.
As with BERT, GPT-4 and GPT-3.5 give more consistent results than LLaMA, Mistral/Mixtral, and Gemini-Ultra.
}

}

\begin{table*}[h!]
\centering
\begin{tabular}{rrrrrrrrrrr}
 &  & QNLI & QQP & RTE & SST2 & WNLI & MNLI & MRPC & COLA %& STSB & Avg 
  \\ \hline
& Average & 265.26 & 279.11 & 11.01 & 170.34 & 5.17 & 627.12 & 12.31 & 25.29 & % &
\\ \hline
%& Classic AL & & & & & & & & &%&&
%\\
& LC & 81.58 & 271.40 & 6.39 & 53.11 & 3.55 & 297.00 & 5.50 & 11.07 &%&&
\\
\rotatebox[origin=c]{90}{AL} & BALD & 812.26 & n.a* & 22.34 & 520.11 & 8.37 & 1610.23 & 30.54 & 68.26 &%&&
\\
& EKM & 85.16 & 289.39 & 7.23 & 53.53 & 4.40 & 306.43 & 6.08 & 10.02 &%&&
\\
& PE & 82.02 & 276.54 & 6.47 & 53.01 & 4.37 & 293.24 & 5.58 & 10.22 &%&&
\\\hline
\end{tabular}
\caption{\revisedNEW{Runtime of the AL strategies (LC, BALD, EKM, and PE) on the GLUE tasks from Table \ref{tab:main_few_shot_results}. The runtime (in minutes) refers to the total time required to select 32 instances over 6 AL iterations. \\ **No result available after 48 hours of execution.}}
\label{tab:al_runtimes}
\end{table*}

\revised{
\subsection{Comparison}
\label{sec:comparison}

\textcolor{gray}{\textit{- How does ActiveLLM compare to other AL strategies?}}

We are interested to see how ActiveLLM performs against other AL methods. 
In Table \ref{tab:main_few_shot_results} we also included the results for the most used AL strategies with transformers identified in our literature overview (see Table \ref{tab:overview}): LC, BALD, EKM, and PE.  
While these methods are implemented with the best preconditions with a model-matching scenario and access to the whole dataset of each task, they are often not able to improve over the baseline and in every case worse than an LLM-variant of ActiveLLM.
This outcome is expected, as these AL strategies face a cold-start problem.
Therefore, in Section \ref{sec:overcoming_coldstart} we test whether ActiveLLM can be used to overcome the cold-start problem of these techniques.
Furthermore, the practical use of these methods is limited, as the selection of examples to be labeled required at least several minutes and, in the case of larger datasets, several hours \revisedNEW{(see Table \ref{tab:al_runtimes})}. 
For instance, with BALD on the QQP dataset, the approach did not yield any result even after 48 hours of execution. 
In contrast, ActiveLLM eliminates the need for model training during the AL process and only considers a small subset of the unlabeled data, enabling query times of just a few seconds while still achieving higher improvements.

\revisedNEW{Standard deviations for the results of Table \ref{tab:main_few_shot_results} are given in the Appendix Table \ref{tab:main_few_shot_results_std}. 
While the values vary across tasks and models, the overall trend indicates that ActiveLLM does not amplify performance instability, and in many cases, can reduce it substantially.}

\subsection{ActiveLLM and Few-Shot Learning}
\label{sec:few_shot_learning}

\textcolor{gray}{\textit{- Can ActiveLLM improve state-of-the-art few-shot learning methods?}}

}
\begin{figure}[ht]
  \centering
  \includegraphics[width=0.47\textwidth]{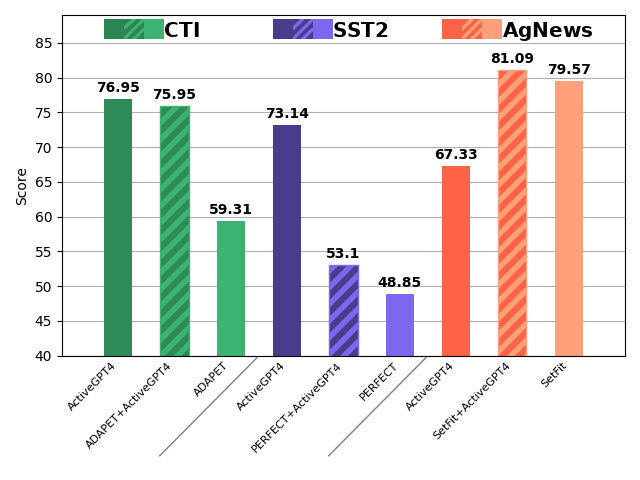}
  \caption{\revised{Comparison of ActiveGPT4, both independently and in combination with few-shot methods, across different tasks \revisedNEW{(32 instances)}. The methods include ADAPET (evaluated using BERT-base on CTI with F1), PERFECT (using RoBERTa-Large on SST2 with accuracy), and SetFit (using BERT-base on AGNews with accuracy).}}
  \label{fig:comparisons}
\end{figure}

\revised{
While showing significant improvements in few-shot scenarios compared to random sampling and other AL strategies, one might ask why not directly use few-shot learning methods.
Therefore, we evaluate the performance of ActiveGPT4 both individually and in combination with the state-of-the-art few-shot learning methods ADAPET, PERFECT, and SetFit.
For each method, we have to use a different dataset for which an implementation is available: ADAPET with CTI, PERFECT with SST-2\footnote{The original implementation only supports RoBERTa-Large.}, and SetFit with AgNews.

The results of this experiment are shown in Figure \ref{fig:comparisons}. 
ActiveGPT4 significantly outperforms the few-shot learning methods ADAPET and PERFECT (+ 17.64\% + 24.29\%).
While not outperforming SetFit on its own, ActiveLLM, when combined with each few-shot learning method, significantly improves all of these methods.
This shows that few-shot learning methods can be improved through the selection of instances, and in some cases, this selection is even more important than the use of few-shot learning methods.

}
\subsection{Non-Few-Shot-Scenarios}
\label{sec:non-few-shot}

In our last experiments, we investigate how ActiveLLM performs in non-few-shot scenarios and if it can be used to overcome the cold-start problem of other AL strategies. 
\revised{For this purpose, we select a sample size of 25 examples per AL iteration and evaluate the process up to 300 examples.}

\subsubsection{Iterated Querying}
\label{sec:iterated_querying}

\textcolor{gray}{\textit{- How does the method perform in non-few-shot-scenarios?}}

\begin{figure}[htp]
  \centering
  \includegraphics[width=0.47\textwidth]{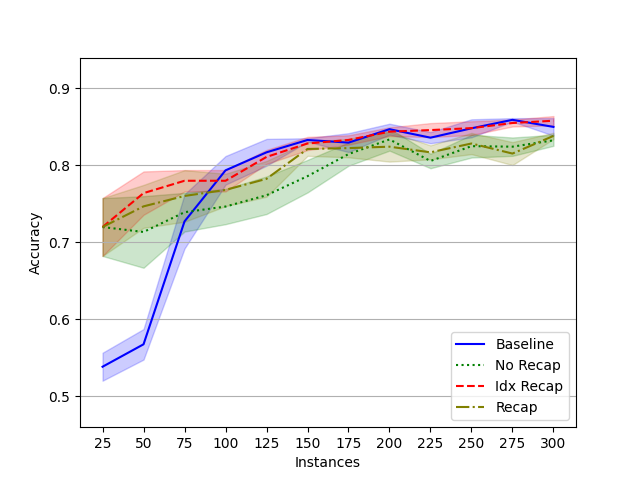}
  \caption{Iterated querying on the SST-2 dataset with different strategies of recalling already labeled instances. All results are averaged over 5 runs (accuracy).} %The standard deviation is shown as tubes around the lines.}
  \label{fig:iterated_querying}
\end{figure}

Figure \ref{fig:iterated_querying} indicates that ActiveGPT4 is particularly beneficial in few-shot scenarios up to about 100 instances. 
\revisedNEW{Beyond this point, it performs comparably to random sampling.}
\revised{Furthermore, the figure highlights the importance of recalling instances that have already been selected, as the no-recap mode performs the worst.} 
Providing the model with the indices of these instances appears to be the most efficient method, as it does not overly complicate or enlarge the prompt.

\revised{These results motivate the subsequent experiment to investigate whether ActiveLLM could be a viable solution to overcome the cold-start problem of learning strategies.}

\subsubsection{Overcoming the Cold-Start Problem}
\label{sec:overcoming_coldstart}

\textcolor{gray}{\textit{- Is ActiveLLM capable of overcoming the cold-start problem in other AL strategies?}}

\begin{figure}[htp]
  \centering
  \includegraphics[width=0.482\textwidth]{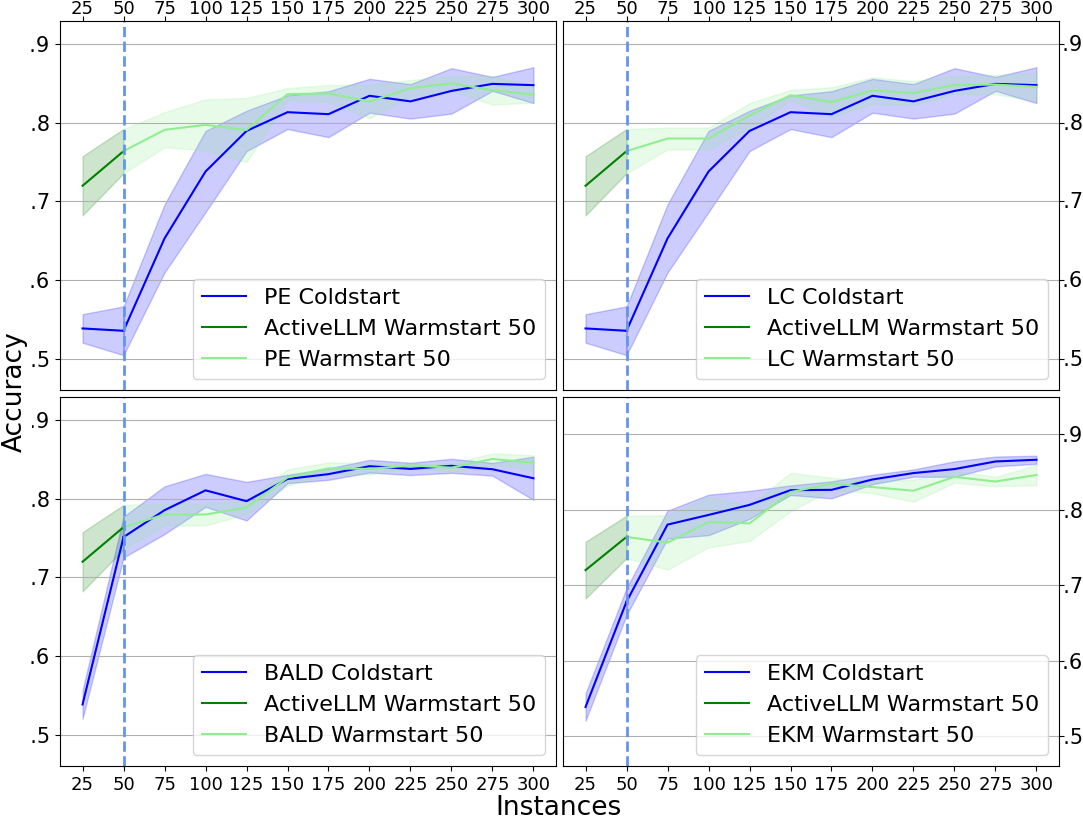}
  \caption{\revised{Comparison of BALD, EKM, LC, and PE with and without selected instances by ActiveGPT4 as seed label set on SST-2. All results are averaged over 5 runs (accuracy).}} %ActiveGPT4 was used to select 50 instances for the SST-2 set with index recap mode and a sample size of 25.  The standard deviation is shown as tubes around the lines.}
  \label{fig:coldstart}
\end{figure}

As outlined in Section \ref{sec:introduction}, a pervasive issue in AL research is the initial poor performance, which leads to random or worse selections.
However, as deep learning models increasingly excel in low-data regimes, the quality of initial data becomes progressively crucial.
Observing that ActiveLLM masters few-shot scenarios, we used the first 50 examples selected by ActiveGPT4 (index recap mode) as the starting point for PE, LC, BALD, and EKM.

%we hypothesized that combining traditional AL strategies with ActiveLLM could be beneficial in non-few-shot scenarios as well. 
%Consequently, in this evaluation, we used the first 50 examples selected by ActiveGPT4 (index recap mode) as the starting point for PE.

Once again, a selection size of 25 was used, and the experiments continued until reaching a total of 300 examples.
\revised{The results in Figure \ref{fig:coldstart} show that ActiveLLM alone outperforms all other methods within the first 50 examples.
Using these examples as a starting point for LC and PE outperforms cold-start strategies, with performance plateauing at around 125 instances.
For EKM and BALD, this effect cannot be reproduced, as the cold-start problem is not as severe with 50 examples as it is for LC and PE.
Nonetheless, the experiment demonstrates that, in cases of a cold-start problem, ActiveLLM is an effective method for addressing it.}

%% file: Sections/05_Conclusion.tex
\section{Discussion \& Conclusion}

Utilizing instruction-tuned LLMs presents several advantages for AL with BERT-like transformers.
Since these LLMs exhibit substantial zero-shot capabilities, they overcome the cold-start problem.
\revised{This is crucial for low-data regimes and few-shot scenarios, which are becoming increasingly important \citep{Yu2020FewShotImportance}.
Realized in the ActiveLLM method, we show that this procedure indeed excels in few-shot scenarios, outperforming the baseline and common AL strategies for transformers.
Furthermore, in our experiments, ActiveLLM significantly outperforms state-of-the-art few-shot learning methods, ADAPET and PERFECT, while also enhancing the performance of SetFit, ADAPET, and PERFECT when used in combination.
We also demonstrate a variant of ActiveLLM with iterative querying, which is applicable to non-few-shot scenarios and incorporates feedback from previous iterations.}
In a combined setting, we show that ActiveLLM overcomes the cold-start problem inherent in conventional AL strategies.

\revised{Moreover, ActiveLLM is particularly suitable for practical scenarios as it effectively eliminates the dependency between the query and the successor model.
\revisedNEW{In our experiments, the AL methods LC, BALD, EKM, and PE required several hours to complete (see Table \ref{tab:al_runtimes}), whereas even the more efficient AL methods commonly discussed in the literature typically require several minutes per querying iteration \citep{yuan2020ColdstartActive}.}
While some of the LLMs used in ActiveLLM require significantly more resources, our approach eliminates the need for model training during the AL \revisedNEW{selection process, as required by many other AL methods}. 
Additionally, only a small portion of the unlabeled data is considered, still resulting in significant improvements while requiring just a few seconds for querying.
Furthermore, the \revisedNEW{current} cost-free availability of these LLMs through chat interfaces eliminates the need for extensive resources or financial investment. 
\revisedNEW{If cost-free availability persists, it democratizes AL}, making it accessible to a wider audience rather than limiting it to hypothetical scenarios that assume unlimited time, resources, or funding.}
%Furthermore, these LLMs are accessible through several chat interfaces, enabling practitioners from all domains without a deep machine learning or programming background to freely implement ActiveLLM. 

\subsection{LLM-based Active Learning Strategies}

The question remains as to how LLMs choose instances. 
\revised{While they clearly explain their approach, including avoiding redundancy, selecting diverse instances, and considering ambiguities (see Figure \ref{fig:input_prompt}), it is nearly impossible to assess whether the models truly perform as claimed.}
\revised{However, LLMs are known for their ability to identify topics and patterns in texts, which suggests that diversity-based sampling, a common approach in other procedures, is a valid method.
Due to this property, they can also identify differences between instances, enabling them to recognize unique, representative, frequent, or highly information-dense examples.}
Similarly, we expect these models to perform well on the tasks themselves, which is why they could be relatively proficient at selecting a balanced set of all classes.
However, these models cannot inspect their own internals, which means they are not performing uncertainty sampling based on these internals.
Related to uncertainty, LLMs might be very well suited for identifying difficult or ambiguous examples.
Occasionally, the models mention that they try to filter out any instances that are anomalies that would teach the successor model incorrect patterns.  
In iterated querying mode, the language models often directly mention strategies regarding the feedback, explaining that for selecting the instances, they incorporate past instances and try to avoid redundancy, but also track very ambiguous instances from which more data might be needed in the successor training.

\subsection{Limitations and Future Work}

In this study, we focus on AL for BERT-like transformer models.
\revisedNEW{While LLMs such as GPT-4 offer greater capabilities, they often require substantial computational resources, incur high API costs, and raise concerns regarding data privacy and control \citep{wang2024comprehensivesurveysmalllanguage, samuel2024BertsGen, Treviso2023EfficientNLP, bosley2023we}.
In contrast, BERT-like models remain an important choice for many practical applications due to their efficiency, open availability, and competitive performance when fine-tuned on domain-specific tasks.
Although ActiveLLM leverages LLMs, their use is limited to the labeling phase and is not required after training.
As such, the resulting BERT-like classifier can be deployed without relying on external APIs or continued access to proprietary LLMs.
Nonetheless, it is important to acknowledge that the integration of LLMs during training may still pose privacy challenges with regard to the data exposed to them.
Our approach therefore offers a trade-off: it combines the strengths of LLMs for data-efficient training while preserving the lightweight and self-contained nature of traditional transformer classifiers at inference time.
It might also be of interest to use the LLM directly as an oracle in our framework, i.e., to have it label the selected instances directly.
However, as this study focuses on the selection of training examples, we chose not to introduce additional potential points of failure.
Instead, we refer readers to the work of \citet{Wang2024HumanLLM, walshe2025automaticlabellingopensourcellms}, who explore the use of LLMs in the annotation process.
We also look forward to applying ActiveLLM to LLMs themselves. 
As reported by \citet{zhang2022}, the selection of training examples for in-context learning can have a significant impact on LLM performance, making ActiveLLM a promising method in this context as well.}

\revisedNEW{While the experiments confirm the strong performance of ActiveLLM's general design, we look forward to further exploring fine-grained LLM pipeline and prompt optimization techniques, including modular and programmable approaches \citep{khattab2024dspy}.}
In particular, the context size is critical for ActiveLLM.
If the prompts get too large, the LLMs tend to forget the task or are not able to reason over the various instances.
\revised{Although some models were reported to be able to handle long context inputs, we found that they were not able to reason over long contexts.} 
This is in line with the findings of \citet{hsieh2024RulerWhatsRealContext} who argue that simple retrieval-based tests, like the common needle-in-a-haystack test, are not suitable for testing long context understanding.
In our results, we noticed differences for tasks with longer instances on average.
\revised{We also hypothesize that the length may be the reason why LLMs sometimes omit certain instructions or information, such as the provided labels in our tests of the iterated querying mode.}
We are interested in considering not only the instances as a length restriction but also the token count in future work.
\revisedNEW{Furthermore, in our experiments we wanted to show that the general prompt optimized in Section \ref{sec:prompt_engineering} is applicable to other models and datasets.
Another approach would be to consider the prompt as adaptive and tune it for each model on the respective development set.
While this introduces a new hyperparameter, we expect it to improve performance in most cases.
Furthermore, optimizing the prompt for GPT-4 may also explain why certain types of LLMs perform worse.}

\revised{A common problem in the field of LLMs is that many established and open benchmarks have leaked into the training data. 
This could bias the results of our work, as a model may already be familiar with the dataset.
While we cannot completely avoid this phenomenon, we have tried to use a less known dataset for prompt engineering, as well as smaller and older models, which are less susceptible to data leakage. 
}

\revisedNEW{When we describe ActiveLLM as much more efficient and less resource-intensive for practitioners than other AL methods, we assume the availability of cost-free chat interfaces or low-cost APIs.
Certainly, the LLMs that ActiveLLM relies on consume considerable resources, even if they are only used for a very short time during sample selection for labeling.
In terms of practicality, our experiments also focus on the use of the chat versions of the LLMs.}
However, directly using the models, e.g., with the APIs, might yield different results. 
While this might only play an insignificant difference for Llama or Mixtral, some chat interfaces, like those from OpenAI or Google, hide parameters, preceding prompts, and exact version declarations. 
We see our results as a baseline that might be improved with different parameters or without a preceding prompt in the future.

%Furthermore, we look forward to works further refining the prompt design.
%For example, the prompt could promote that the language model should favor certain AL strategies.
%It would also be interesting to analyze certain strategies in isolation, like only diversity sampling.
%While we did not receive better results incorporating the true labels during iterated querying, this behavior might stem from the fact that we only evaluated ActiveLLM on at most three-class multi-classification tasks.
%We expect with a growing number of classes that the models identify to also look for the count of already labeled instances of a certain class.

%% file: Sections/06_Appendix.tex
\twocolumn

\section*{Appendix} \label{sec:appendix}
\addcontentsline{toc}{section}{Appendices}
\renewcommand{\thesubsection}{\Alph{subsection}}
\setcounter{subsection}{0}

\subsection{Full Prompt Design for Iterated Querying Mode}
\label{sec:prompt_design_feedback_mode}

\begin{figure*}[htbp]
  \centering
  \includegraphics[width=0.7\textwidth]{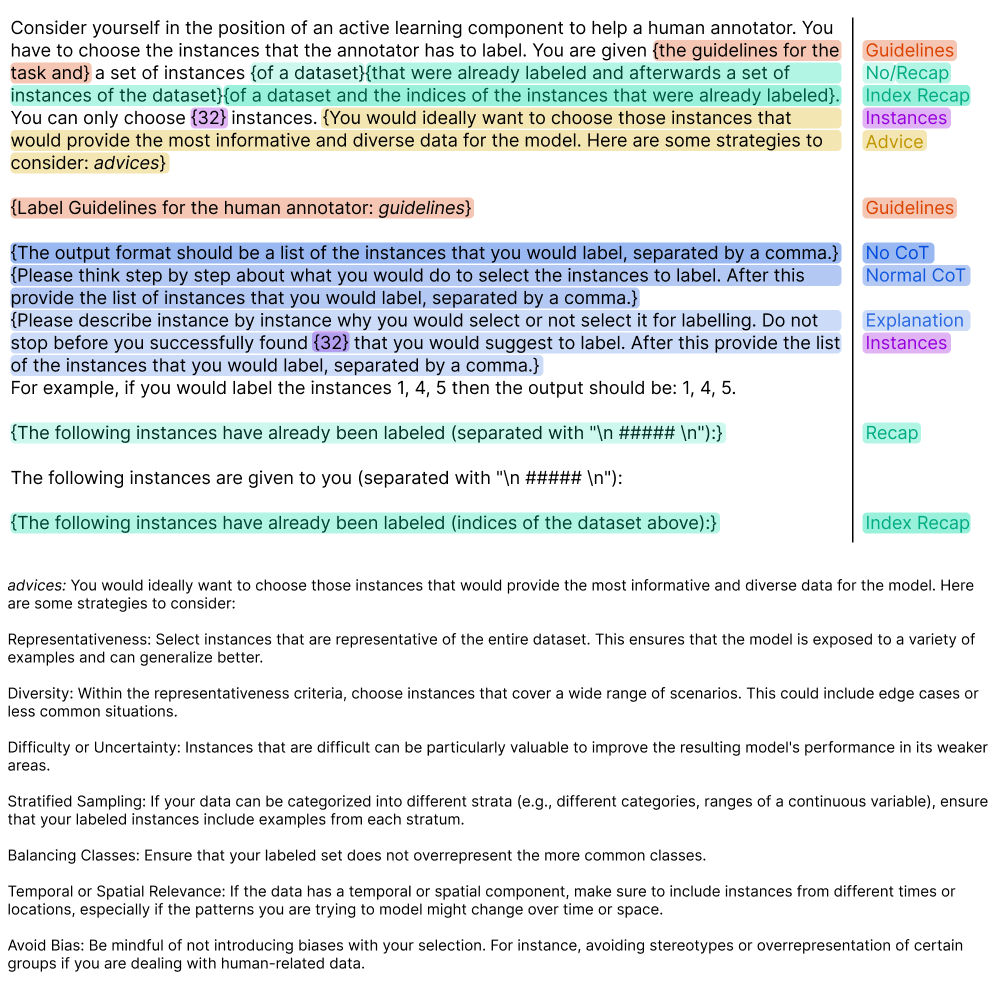}
  \caption{\revised{Complete prompt design for the iterated querying mode of ActiveLLM, including the text of the 'advice' parameter, which represents the description of several AL strategies.}}
  \label{fig:prompt_engineering_extended}
\end{figure*}

Figure \ref{fig:prompt_engineering_extended} illustrates the comprehensive prompt design for the second variant of ActiveLLM. 
Compared to the prompt design of the first variant, this one contains the parameters 'No Recap', 'Recap' and 'Index Recap'.
In scenarios 'No Recap' and 'Recap', the unlabeled instances presented to the model are the ones immediately following the last labeled instance, extending up to the index of this last labeled instance plus the batch size, typically 200 instances. 
In contrast, the 'Index Recap' scenario maintains the last presented instance but includes all preceding instances, allowing the model to access texts from the specific indices given as labeled. 
Additionally, Figure \ref{fig:prompt_engineering_extended} shows the 'advice' parameter, which outlines various AL strategies, including representativeness, diversity, difficulty, stratification, balance, temporal/special distribution, and bias avoidance.

\subsection{Specialized CTI Dataset Details}
\label{sec:dataset_appendix}

\begin{table}[h]
    \centering
    \begin{tabular}{l@{\hspace{1em}}l@{\hspace{1em}}l@{\hspace{1em}}l@{\hspace{1em}}}
    \toprule
         \textbf{Split} & \textbf{Count} & \textbf{Relevant} & \textbf{Not Relevant}  \\
         \midrule
         Train (full) & 1800 & 949 & 851 \\
         \hline
         Train & 32 & 16 & 16 \\
         \hline
         Dev (full) & 600 & 273 & 327 \\
         \hline
         Dev & 8 & 4 & 4 \\
         \hline
         Test & 601 & 304 & 297 \\
         \bottomrule
         \textbf{Total} & 3001 & 1526 & 1475 \\
    \bottomrule
    \end{tabular}
    \caption{\revised{Split of the CTI dataset with count of relevant and not relevant labels in the datasets.}}
    \label{tab:dataset_splits}
\end{table}
The Spezialised CTI few-shot dataset consists of tweets that are binary labeled based on their relevance to cybersecurity experts during the Microsoft Exchange Server data breach in 2021.
\revisedNEW{The dataset is split into a full and a few-shot training and development set, with only the full training set being relevant in our case, from which 32 instances are sampled.}
The splits (train, dev) consist of 1800 and 600 instances for the full set and 32 and 8 instances for the few-shot set, respectively. 
The test set is the same in both cases and consists of 601 instances.
The details are also given in Table \ref{tab:dataset_splits}.

\subsection{LLM Details}
\label{sec:llm_details}

For the LLMs GPT-4\footnote{\url{https://chatgpt.com/?model=gpt-4} - queried in January 2024}, GPT-4o\footnote{\url{https://chatgpt.com/?model=gpt-4o} - queried in May 2024}, o1\footnote{\url{https://chatgpt.com/?model=o1} - queried in January 2025}, GPT-3.5\footnote{\url{https://chatgpt.com/?model=text-davinci-002-render-sha} - queried in April 2024}, Gemini-Ultra\footnote{\url{https://gemini.google.com/app} -> Gemini-Ultra - queried in April 2024}, and Mistral Large\footnote{\url{https://chat.mistral.ai/chat} -> Model: Large - queried in April 2024}, we utilized the native chat interfaces provided by OpenAI, Google, and Mistral. For Llama 3 70B\footnote{\url{https://deepinfra.com/meta-llama/Meta-Llama-3-70B-Instruct} - queried in May 2024} and  Mixtral 8x7B\footnote{\url{https://deepinfra.com/mistralai/Mixtral-8x7B-Instruct-v0.1} - queried in April 2024}, we used the chat interface provided by deepinfra. 
We used no extra system prompts, a maximum answer length of 2048 tokens, a temperature of 0.7, Top P of 0.9, and Top K set to 0. 
For all interactions, we opened a new chat session and disabled memory functions.

\begin{figure}[ht]
  \centering
  \includegraphics[width=0.47\textwidth]{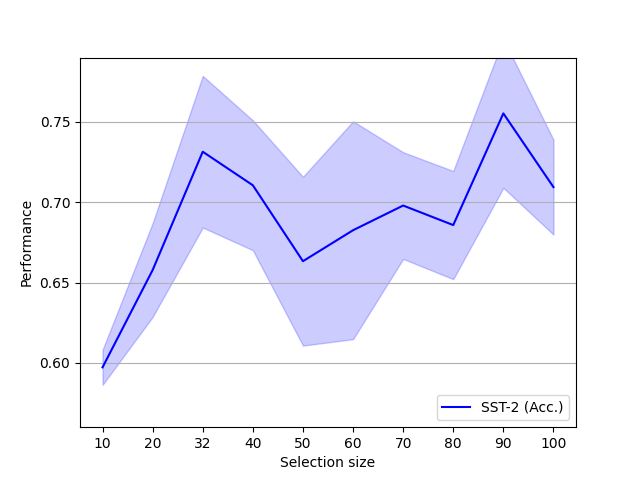}
  \caption{\revised{Varying selection size of examples to be selected by ActiveGPT4. All results are averaged over 5 runs on the SST-2 (accuracy) dataset.}}
  \label{fig:selection_size_sst2}
\end{figure}

%Standard Derivation table%
\begin{table*}[ht!]
\centering
\begin{tabular}{rrrrrrrrrr}
 &  & QNLI & QQP & RTE & SST2 & WNLI & MNLI-(m/mm) & MRPC & COLA
  \\ \hline
& Baseline & 11.10 & 1.08 & 1.31 & 2.09 & 3.54 & 2.43/3.23 & 2.46 & 8.32
\\ \hline
& GPT-4 & 9.53 & 2.02 & 3.61 & 5.28 & 1.61 & 2.20/2.77 & 2.48 & 3.45
\\
\rotatebox[origin=c]{90}{200} & GPT-4o & 10.70 & 0.60 & 1.90 & 5.28 & 2.89 & 0.33/0.16 & 1.89 & 4.09
\\
& o1 & 8.52 & 2.41 & 2.25 & 2.06 & 2.82 & 2.18/1.23 & 0.49 & 2.25
\\
& Mistral Large & 10.77 & 1.85 & 2.41 & 4.46 & 1.61 & 1.69/1.23 & 1.42 & 1.87
\\\hline
 & Llama 3 70B & 10.34 & 1.64 & 2.06 & 1.87 & 1.89 & 0.86/1.04 & 2.87 & 2.61
\\
& GPT-3.5 & 11.10 & 1.66 & 1.99 & 5.43 & 5.33 & 2.93/3.27 & 2.37 & 2.26
\\
\rotatebox[origin=c]{90}{100} & Mistral Large & 9.11 & 1.42 & 1.96 & 4.48 & 2.23 & 2.55/3.08 & 0.80 & 1.34
\\
& Gemini-Ultra  & 11.65 & 3.68 & 1.34 & 0.00 & 1.61 & 2.40/2.79 & 0.62 & 0.24
\\
& Mixtral 8x7B  & 4.24 & 2.05 & 1.29 & 2.55 & 3.67 & 2.26/3.18 & 0.55 & 1.74
\\
\hline
%& Classic AL & & & & & & & & &%&&
& LC & 12.10 & 1.43 &1.81 & 2.28 & 2.14 & 3.18/3.38 & 2.67 & 1.41
\\
\rotatebox[origin=c]{90}{AL} & BALD & 9.05 & n.a. & 0.66 & 1.59 & 3.21 & 3.23/3.75 & 0.13 & 6.97
\\
& EKM & 4.90 & 4.59 & 0.96 & 2.35 & 1.00 & 0.66/0.70 & 5.35 & 3.63
\\
& PE & 9.20 & 1.43 & 1.81 & 2.28 & 2.14 & 2.29/2.65 & 2.67 & 1.41
\\\hline
\end{tabular}
\caption{\revised{Standard deviations of the results in Table \ref{tab:main_few_shot_results}.}}
\label{tab:main_few_shot_results_std}
\end{table*}

\begin{table*}[ht]
{\setlength{\tabcolsep}{4pt}
\centering
%\hspace*{0.1em}-\hspace*{0.1em}  \textcolor{red}{$\downarrow$} $\uparrow$%
\begin{tabular}{llllllllll}
 &  & QNLI & QQP & RTE & SST2 & WNLI & MNLI-(m/mm) & MRPC & COLA
  \\ \hline
& Baseline & 60.46 & 66.50 & 50.69 & 58.39 & 33.24 & 33.76/33.32 & 73.63 & 69.03
\\ \hline
& GPT-4  & \textbf{67.63} $\uparrow$ & 65.30 \textcolor{red}{$\downarrow$} & 54.44 $\uparrow$ & \textbf{82.00} $\uparrow$ & 42.54 $\uparrow$ & 34.46/34.14 $\uparrow$ & 70.39 \textcolor{red}{$\downarrow$} & 69.20 $\uparrow$
\\
\rotatebox[origin=c]{90}{200} & GPT-4o  & 65.54 $\uparrow$ & 63.57 \textcolor{red}{$\downarrow$} & 53.21 $\uparrow$ & 68.78 $\uparrow$ & 43.10 $\uparrow$ & 32.49/32.99 \textcolor{red}{$\downarrow$} & \textbf{74.46} $\uparrow$ & 69.20 $\uparrow$
\\
& o1 & 57.57 \textcolor{red}{$\downarrow$} & \textbf{67.71} $\uparrow$ & 58.63  $\uparrow$& 68.65 $\uparrow$ & 42.54 $\uparrow$ & 34.45/33.78 $\uparrow$ & 68.38 \textcolor{red}{$\downarrow$} & 69.19 $\uparrow$
\\
& Mistral Large  & 58.98 \textcolor{red}{$\downarrow$} & 66.63 $\uparrow$ & 57.33 $\uparrow$ & 61.56 $\uparrow$ & 54.37 $\uparrow$ & 35.59/35.33 $\uparrow$ & 69.12 \textcolor{red}{$\downarrow$} & 69.19 $\uparrow$ 
\\\hline
 & Llama 3 70B  & 58.44 \textcolor{red}{$\downarrow$} & 65.46 \textcolor{red}{$\downarrow$} & 53.14 $\uparrow$ & 77.39 $\uparrow$ & 43.38 $\uparrow$ & 36.03/35.51 $\uparrow$ & 68.63  \textcolor{red}{$\downarrow$} & 69.34 $\uparrow$
\\
& GPT-3.5  & 60.47 $\uparrow$ & 67.21 $\uparrow$ & 55.81 $\uparrow$ & 68.39 $\uparrow$ & \textbf{58.31} $\uparrow$ & 35.56/36.22 $\uparrow$ & 72.01 \textcolor{red}{$\downarrow$} & 69.13 $\uparrow$
\\
\rotatebox[origin=c]{90}{100} & Mistral Large  & 59.36  \textcolor{red}{$\downarrow$} & 66.91 $\uparrow$ & 49.96 \textcolor{red}{$\downarrow$} & 79.66 $\uparrow$ & 50.70 $\uparrow$ & \textbf{36.56/36.03} $\uparrow$ & 68.38 \textcolor{red}{$\downarrow$} & 69.20 $\uparrow$
\\
& Gemini-Ultra  & 60.30 \textcolor{red}{$\downarrow$} & 66.23 \textcolor{red}{$\downarrow$} & 54.37 $\uparrow$ & 49.08 \textcolor{red}{$\downarrow$} & 40.00 $\uparrow$ & 35.55/36.20 $\uparrow$ & 69.95 \textcolor{red}{$\downarrow$} & \textbf{69.38} $\uparrow$
\\
& Mixtral 8x7B  & 61.15 $\uparrow$ & 66.30 \textcolor{red}{$\downarrow$} & \textbf{58.77} $\uparrow$ & 71.97 $\uparrow$ & 42.25 $\uparrow$ & 33.31/33.46 \textcolor{red}{$\downarrow$} & 67.70 \textcolor{red}{$\downarrow$} & 69.12 $\uparrow$
\\
\hline
\end{tabular}}
\caption{\revised{RoBERTa: Few-shot results (32 instances) of ActiveLLM with different LLMs on GLUE tasks. `Active 200/100' describes the size of the batch of unlabeled instances. All results are averaged over 5 runs (accuracy).}} %Only GPT-4, GPT-4o and Mistral Large are capable of processing 200 instances. All results are averaged over 5 runs (accuracy).}
\label{tab:main_few_shot_results_roberta}
\end{table*}

\begin{figure*}[t] 
\centering
  \includegraphics[width=1\linewidth]{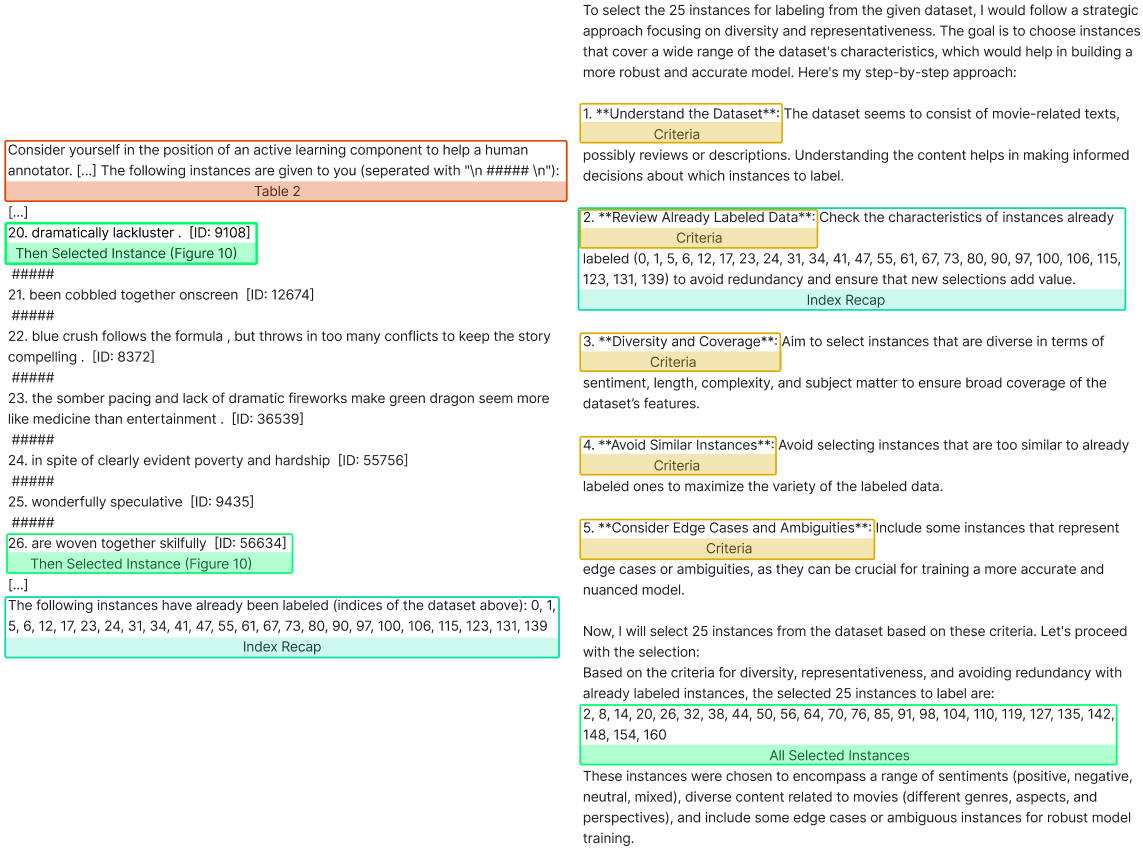}
  \caption{\revised{Example prompt for SST-2 generated by ActiveLLM in iterated querying mode, no guidelines, and no advice. (left) \\
  GPT-4 response to the prompt shown on the left. (right)}}
  \label{fig:input_prompt}
\end{figure*}